\documentclass[twocolumn]{article}

\usepackage{cvpr}
\usepackage{times}
\usepackage{epsfig}
\usepackage{graphicx}
\usepackage{amsmath}
\usepackage{amssymb}
\usepackage{booktabs}
\usepackage[pagebackref=true,breaklinks=true,letterpaper=true,colorlinks=true,citecolor=blue,bookmarks=false]{hyperref}

\cvprfinalcopy % *** Uncomment this line for the final submission

\begin{document}

%%%%%%%%% TITLE
\title{Feature Space Transfer for Data Augmentation}

\author{Bo Liu\\
University of California, San Diego\\
{\tt\small boliu@ucsd.edu}
\and
Xudong Wamg\\
University of California, San Diego\\
{\tt\small xuw080@ucsd.edu}
\and
Mandar Dixit\\
Microsoft\\
{\tt\small madixit@microsoft.com}
\and
Roland Kwitt\\
University of Salzburg, Austria\\
{\tt\small rkwitt@gmx.at}
\and
Nuno Vasconcelos\\
University of California, San Diego\\
{\tt\small nuno@ece.ucsd.edu}
}

\maketitle
%\thispagestyle{empty}

%%%%%%%%% ABSTRACT
\begin{abstract}
The problem of data augmentation in feature space is considered. 
A new architecture, denoted the FeATure TransfEr Network (FATTEN),
is proposed for the modeling of feature trajectories induced by variations 
of object pose. This architecture exploits a parametrization of
the pose manifold in terms of pose and appearance. This leads to a 
deep encoder/decoder network architecture, where the encoder
factors into an appearance and a pose predictor. Unlike previous
attempts at trajectory transfer, FATTEN can be efficiently trained 
end-to-end, with no need to train separate feature transfer functions.
This is realized by supplying the decoder with information about 
a target pose and the use of a multi-task loss that penalizes
category- and pose-mismatches. In result, FATTEN discourages discontinuous 
or non-smooth trajectories that fail to capture the structure of the 
pose manifold, and generalizes well on object recognition tasks involving
large pose variation. Experimental results on the artificial
ModelNet database show that it can successfully learn to map source 
features to target features of a desired pose, while preserving class identity.
Most notably, by using feature space transfer for data augmentation 
(w.r.t. pose and depth) on SUN-RGBD objects, we demonstrate considerable 
performance improvements on one/few-shot object recognition in a transfer
learning setup, compared to current state-of-the-art methods.
\end{abstract}

%%%%%%%%% BODY TEXT
\section{Introduction}

Convolutional neural networks (CNNs) trained on large datasets, such
as ImageNet \cite{Deng09a}, have enabled tremendous gains in problems like object
recognition over the last few years. These models not only achieve
human level performance in recognition challenges, but are also
easily transferable to other tasks, by fine tuning. Many recent works
have shown that ImageNet trained CNNs, like AlexNet~\cite{Krizhevsky12a},
VGG~\cite{Simonyan14a}, GoogLeNet~\cite{Szegedy15a}, or ResNet~\cite{He16a}
can be used as feature extractors for the solution of problems as diverse
as object detection \cite{Girshick15a,Ren15a} or generating image 
captions~\cite{Karphaty15a,Vinyals15a}. 
Nevertheless, there are still challenges to CNN-based recognition.
One limitation is that existing CNNs still have limited ability to handle
pose variability. This is, in part, due to limitations of existing
datasets, which are usually collected on the web and are biased towards
a certain type of images. For example, objects that have a well
defined ``frontal view,'' such as ``couch'' or ``clock,'' are rarely
available from viewing angles that differ significantly from frontal.

%\begin{figure}[h!]
%\includegraphics[width=\columnwidth]{figs/Intro}
%\caption{Schematic illustration of \emph{feature space transfer} for
%variations in \emph{pose}. The
%\textcolor{red}{inputs} to feature space transfer are (1) a CNN activation 
%$\mathbf{x}$ of some object
%(here: airplane) and (2) the desired attribute value, encoded as a 1-hot 
%vector (here: $[0,\mathbf{1},0,0]$, corresponding to a $30^\circ$ rotation \wrt
%the $z$-axis) of the partitioned attribute range. Output is a \emph{synthetic}
%CNN activation $\hat{\mathbf{x}}$ corresponding to the desired change 
%along the feature trajectory.}
%\end{figure}

\begin{figure}[t!]
	\centering
	\includegraphics[width=0.85\columnwidth]{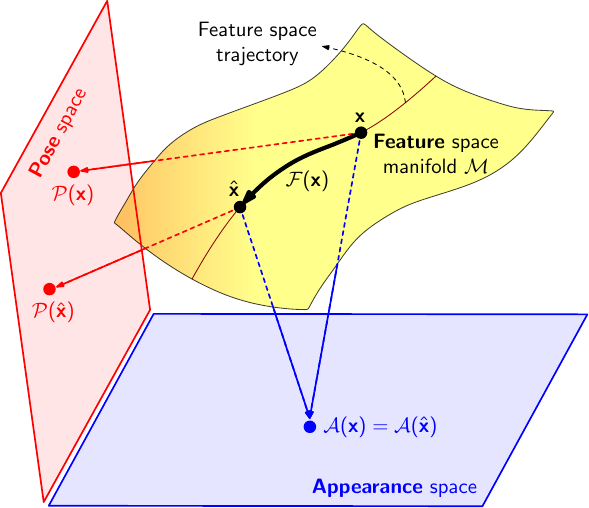}
	\caption{Schematic illustration of \emph{feature space transfer} for
		variations in \emph{pose}. The input feature $\mathbf{x}$ and transferred feature $\mathbf{\hat{x}}$ are projected to the same point in \emph{appearance space}, but have different mapping points in \emph{pose space}.\label{fig:intro}}
\end{figure}

\vskip0.5ex
This is problematic for applications like robotics, where a robot
might have to navigate around or manipulate such objects. When
implemented in real time, current CNNs tend to produce object
labels that are unstable with respect to viewing angle. The resulting
object recognition can vary from nearly perfect under some views to
much weaker for neighboring, and very similar, views. One potential
solution to the problem is to rely on larger datasets, with a much
more dense sampling of the viewing sphere. This, however, is not
trivial to accomplish for a number of reasons. \emph{First}, for many classes,
such images are not easy to find on the web in large enough
quantities. \emph{Second}, because existing recognition methods are weakest at
recognizing ``off-view'' images, the process cannot be easily
automated. \emph{Third}, the alternative of collecting these images in the
lab is quite daunting. While this has been done in the past, e.g.\,
the COIL~\cite{Nene96a}, NORB~\cite{LeCun04a}, or Yale face dataset,
these datasets are too small by modern standards. The set-ups
used to collect them, by either using a robotic table
and several cameras, or building a camera dome, can also not be easily
replicated and do not lend themselves to distributed dataset creation
efforts, such as crowd sourcing. Finally, even if feasible to assemble,
such datasets would be massive and thus difficult to process. 
For example, the NORB recommendation of collecting $9$ elevations, $36$
azimuths, and $6$ lighting conditions per object, results in $1,944$
images per object. Applying this standard to ImageNet would result
in a dataset of close to $2$ billion images!

\vskip0.5ex
Some of these problems can be addressed by resorting to computer generated
images. This has indeed become an established practice to address
problems that require multiple object views, such as shape recognition,
where synthetic image datasets~\cite{Peng15a,Su15a} are routinely used.
However, the application of networks trained on synthetic data to
real images raises a problem of \emph{transfer learning}. 
While there is a vast literature literature on this 
topic~\cite{Socher13a,Lampert14a,RomeraParedes15a,Tang10a,Vinyals16a,
Santoro16a,Ravi17a}, these methods are usually not tailored for 
the transfer of object poses. In particular, they do not explicitly account
for the fact that, as illustrated in Fig.~\ref{fig:intro}, objects subject 
to pose variation span low-dimensional manifolds of image space, 
or corresponding spaces of CNN features. This has recently been addressed
by~\cite{Dixit17a}, who have proposed an attribute guided augmentation (AGA) 
method to transfer object trajectories along the pose manifold. 

Besides learning a classifier that generalizes on target data, the
AGA transfer learning system also includes a module that predicts
the responses of the model across views. More precisely,
given a view of an unseen object, it predicts the model responses
to a set of other views of this object. These can then be used to
augment the training set of a one-shot classifier, \ie, a classifier 
that requires a single image per object for training.
While this was shown to improve on generic transfer learning methods, 
%, due to the fact that it does not model
%continuous trajectories. Instead, 
AGA has some limitations. For example, it discretizes the pose angle into
several bins and learns an independent trajectory transfer function between each possible pair of them.
While this simplifies learning, the trajectories are not
guaranteed to be continuous. Hence, the modeling fails to capture some of the
core properties of the pose manifold, such as continuity and smoothness.
In fact, a $360^\circ$ walk around the viewing sphere is not
guaranteed to have identical start and finishing feature responses.
In our experience, these choices compromise the effectiveness of the transfer.

\vskip0.5ex
\noindent
\textbf{Contribution.}
In this work, we propose an alternative, \emph{FeATure TransfEr 
Network (FATTEN)}, that addresses these problems. 
Essentially, this is an encoder-decoder 
architecture, inspired by Fig.~\ref{fig:intro}. We exploit a
parametrization of pose trajectories in terms of an \emph{appearance
map,} which captures properties such as object color and texture and
is constant for each object, and a \emph{pose map,} which is pose 
dependent. The \emph{encoder} maps the feature responses $\bf x$ of a CNN
for an object image into a pair of appearance ${\cal A}({\bf x})$
and pose ${\cal P}({\bf x})$ parameters. The \emph{decoder} then takes
these parameters plus a \emph{target} pose ${\bf t} = {\cal P}(\hat{\bf x})$
and produces the corresponding feature vector $\hat{\bf x}$. The network 
is trained end-to-end, using a multi-task loss that accounts for both 
classification errors and the accuracy of feature transfer across views.

The performance of FATTEN is investigated on two tasks. The first
is a \emph{multi-view retrieval} task, where synthesized 
feature vectors are used to retrieve images by object class and pose.
These experiments are conducted on the popular ModelNet~\cite{Wu15a}
shape dataset and show that FATTEN generates features of good
quality for applications involving computer graphics imagery. This
could be of use for a now large 3D shape classification 
literature~\cite{Wu15a,Qi16a,su2015multi,Qi16b},
where such datasets are predominant. The \emph{second} task is
transfer learning. We compare the performance of the
proposed architecture against both general purpose transfer learning
algorithms and the AGA procedure. Our results show that there are
significant benefits in developing methods explicitly for trajectory
transfer, and in forcing these methods to learn continuous
trajectories in the pose manifold. The FATTEN architecture is shown
to achieve state-of-the-art performance for pose transfer. 

% The \emph{second} task
% is multi-view recognition. This addresses the fact that the collection
% of multi-view datasets is a challenging problem. Even when an initial
% dataset, collected in the lab, is available it can be difficult to augment it
% with new objects. We investigate whether the ability to transfer pose
% trajectories can be used to reduce the number of views that must
% be collected per object. In the absence of large multi-view datasets of
% real objects, we perform experiments on the synthetic datasets widely
% used for shape recognition~\cite{Wu15a} Our results show that, at least
% in this realm, it is possible to significantly reduce the number of views
% that are needed to achieve a certain performance level by using
% trajectory transfer.

\vskip1ex
\noindent
\textbf{Organization.} In Sect.~\ref{section:relatedwork}, we review
related work; Sect.~\ref{section:fst} introduces the proposed FATTEN
architecture. Sect.~\ref{section:experiments} presents experimental results on 
ModelNet and SUN-RGBD and Sect.~\ref{section:discussion} concludes
the paper with a discussion of the main points and an outlook on
open issues.

%------------------------------------------------------------------------
\section{Related Work}
\label{section:relatedwork}

Since objects describe smooth trajectories in image space,
as a function of viewing angle, it has long been known that
such trajectories span a 3D manifold in image space,
parameterized by the viewing angle. Hence, many of the
manifold modeling methods proposed in the literature~\cite{Roweis00a,Belkin03a,vdM08a} could, in 
principle, be used to develop
trajectory transfer algorithms. However, many of these
methods are transductive, \ie, they do not produce a function
that can make predictions for images outside of the training set,
and do not leverage recent advances in deep learning.
While deep learning could be used to explicitly
model pose manifolds, it is difficult to rely on CNNs
pre-trained on ImageNet for this purpose.
This is because these networks attempt to collapse the manifold
into a space where class discrimination is linear. On the other hand,
the feature trajectories in response to pose variability
are readily available. These trajectories are also much easier to
model. For example, if the CNN is successful in mapping the
pose manifold of a given object into a single point, \ie,
exhibits total pose invariance for that object, 
the problem is already solved and trajectory leaning is trivial for
that object.

\vskip0.5ex
One of the main goals of trajectory transfer is to ``fatten'' a
feature space, by augmenting a dataset with feature responses of
unseen object poses. In this sense, the problem is related to extensive 
recent literature on GANs~\cite{Goodfellow14a},
 which have been successfully used to generate 
images, image-to-image translations~\cite{Isola17a}, 
%attributes~\cite{...} 
inpainting \cite{Pathak16a} or style-transfer~\cite{Gatys16a}. While 
our work uses an encoder-decoder
architecture, which is fairly common in the GAN-based image generation
literature, we aim for a different goal of generating CNN feature
responses. This prevents access to a dataset of ``real'' feature responses
across the pose manifold, since these are generally unknown. While an ImageNet
CNN could be used to produce some features, the problem that we are
trying to solve is exactly the fact that ImageNet CNNs do not
effectively model the pose manifold. Hence, the GAN formalism of
learning to match a ``real'' distribution is not easily applicable to
trajectory transfer.

Instead, trajectory transfer is more closely related to the topic of transfer learning, where, now, there is extensive work on problems such as zero-shot~\cite{Socher13a,Lampert14a,RomeraParedes15a} or 
$n$-shot~\cite{Tang10a,Vinyals16a,Santoro16a,Ravi17a} learning.
%literature on problems such as 
%zero-shot~\cite{Socher13a,Lampert14a,RomeraParedes15a} or 
%$n$-shot~\cite{Tang10a,Vinyals16a,Santoro16a,Ravi17a} learning.
However, these
methods tend to be of general purpose. In some cases, they
exploit generic semantic properties, such as attributes or
affordances~\cite{Lampert14a,RomeraParedes15a}, in others they simply rely on generic
machine learning for domain adaptation~\cite{Socher13a}, 
transfer learning~\cite{Vinyals16a} or, more recently, 
meta-learning \cite{Santoro16a,Flin17a,Ravi17a}. None of these methods
exploits specific properties of the pose manifold, such as the
parametrizations of Figure~\ref{fig:intro}. The introduction of
networks that enforce such parameterizations is a form of
regularization that improves on the transfer performance of
generic procedures. This was shown on the AGA work~\cite{Dixit17a} and
is confirmed by our results, which show even larger gains over
very recent generic methods, such as feature hallucination proposed in~\cite{hariharan2016low}.

\vskip0.5ex
Finally, trajectory transfer is of interest for problems involving
multi-view recognition. Due to the increased cost of multi-view
imaging, these problems frequently include some degree of learning
from  computer generated images. This is, for example, an established 
practice in the shape recognition literature,
%Some of these problems can be addressed by resorting to computer generated
%images. This has indeed become an established practice to address
%problems that require multiple object views, such as shape recognition,
where synthetic image datasets~\cite{Peng15a,Su15a} are routinely used.
The emergence of these artificial datasets has enabled
a rich literature in shape recognition methods~\cite{knopp2010hough, Wu15a, su2015multi, Qi16a, Qi16b, Kalogerakis16a} and
already produced some interesting conclusions. For example,
while many representations have been proposed, there is some evidence that
the problem could be solved as one of multi-view recognition, using
simple multi-view extensions of current CNNs \cite{su2015multi}. It is
not clear, however, how these methods or conclusions generalize to real
world images. Our results show that feature trajectory transfer models, 
such as FATTEN, learned on synthetic datasets, such as ModelNet~\cite{Wu15a}, 
can be successfully transferred to real image datasets, such as 
SUN-RGBD~\cite{Song15a}.

% One possibility is to rely on \emph{transfer learning}. Several
% works have shown that it is possible to leverage large datasets
% of synthetic examples to help solve real tasks~\cite{adversarialimpostersrefs}.
% However, these methods were mostly developed before the introduction
% of large CNNs, improving the generalization of recognition methods
% significantly more prone to dataset bias~\cite{...}. It is not clear
% how effective they would be when applied to the much more robust
% CNN models currently used today.

\section{The FATTEN architecture}
\label{section:fst}

In this section, we describe the proposed architecture
for \emph{feature space transfer}. 

\subsection{Motivation}

In this work, we assume the availability of a training set with pose
annotations, \ie, $\{(\mathbf{x}_n, {\bf p}_n, y_n)\}_n$, where 
$\mathbf{x}_n \in \mathbb{R}^D$ is the feature vector (\eg, a CNN 
activation at some layer) extracted from an image, ${\bf p}_n$ is the 
corresponding pose value and $y_n$ a category label. 
The pose value could be a scalar $p_n$, \eg, the azimuth 
angle on the viewing sphere, but is more generally a vector,
\eg, also encoding an elevation angle or even the distance to the object
(object depth). The problem is to learn the feature transfer 
function ${\cal F}(\mathbf{x}_n, {\bf p})$ that maps the source feature 
vector $\mathbf{x}_n$ to a target feature vector $\hat{\mathbf{x}}_n$ 
corresponding to a new pose $\bf p$. 
% With an external training data set with rich attributes $\{(\mathbf{x}_n, t_n, y_n)\}_n$, 
% where $\mathbf{x}_n$ is the feature of a given sample in the data set, $t_n$ is the corresponding attributes, and $y_n$ is the category label. 

\subsection{The FATTEN Architecture}

The FATTEN architecture is inspired by Fig.~\ref{fig:intro},
which depicts the manifold spanned by an object under pose
variation. The manifold ${\cal M}$ is embedded in $\mathbb{R}^D$ and 
is parameterized by two variables. The first, is an {\it appearance 
descriptor\/} $\mathbf{a} \in \mathbb{R}^A$ that captures object 
properties such as color or texture. This parameter is pose
invariant, \ie, it has the same value for all points on the manifold.
It can be thought of as an object identifier that distinguishes the
manifold spanned by one object from those spanned by others.
The second is a {\it pose descriptor\/} $\mathbf{p} \in \mathbb{R}^N$
that characterizes the point ${\bf x}$ on the manifold that corresponds
to a particular pose $\mathbf{p}$. Conceptually, feature points 
$\mathbf{x}$ 
could be thought of as the realization of a mapping
\begin{equation}
\phi(\mathbf{a},\mathbf{p}) \mapsto \mathbf{x} \in \mathcal{M}\enspace.
\label{eqn:manifoldmapping}
\end{equation}
%The manifold $\cal M$ is thus
%the range space of the mapping 
%\begin{equation}
%  {\bf x} = {\cal M}({\bf a},{\bf p}).
%  \label{fig:manifold}
%\end{equation}

\begin{figure}[t!]
\centering
\includegraphics[scale=0.92]{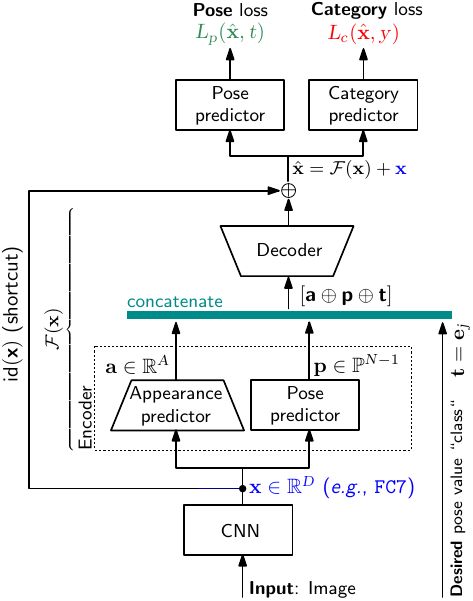}
\caption{The FATTEN architecture. Here, $\text{id}$ denotes
  the identity shortcut connection, $D$ the dimensionality of the input 
  feature space, $C$ the dimensionality of the appearance 
  space and $\mathbb{P}^{N-1}$ the $N-1$ probability simplex.
\label{fig:arch}}
\end{figure}

The FATTEN architecture models the relationship between the
feature vectors extracted from object images and the associated 
appearance and pose parameters. As shown in Fig.~\ref{fig:arch}, it is an
encoder/decoder architecture. The encoder essentially aims to invert the 
mapping of~\eqref{eqn:manifoldmapping}. Given
a feature vector $\bf x$, it produces an estimate of the 
appearance $\bf a$ and pose $\bf p$ parameters. This is complemented
with a {\it target\/} pose parameter $\bf t$, which specifies the
pose associated with a desired feature vector $\hat{\bf x}$.
This feature is then generated by a decoder that that operates
on the concatenation of $\mathbf{a}$, $\mathbf{p}$ and $\mathbf{t}$, 
\ie, $[{\bf a}, {\bf p}, {\bf t}]$.
 While, in principle,
it would suffice to rely on $\hat{\bf x} = {\phi}({\bf a},{\bf t})$, 
\ie, to use the inverse of the encoder as a decoder, we have 
obtained best results with the following modifications.

\vskip0.5ex
\emph{First}, to discourage the encoder/decoder pair from learning a mapping
that simply ``matches'' feature pairs, FATTEN implements the residual 
learning paradigm of \cite{He16a}. In particular,
the encoder-decoder is only used to learn the residual 
\begin{equation}
  {\cal F}({\bf x}) = \hat{\bf x} - {\bf x}
\end{equation}
between the target and source feature vectors. \emph{Second}, two mappings that 
explicitly recover the appearance $\bf a$ and pose $\bf p$ are used
instead of a single monolithic encoder. This facilitates
learning, since the pose predictor can be learned with full
supervision. Third, a vector encoding is used for the source $\bf p$ and
target $\bf t$ parameters, instead of continuous values. This 
makes the dimensionality of the pose parameters closer to that
of the appearance parameter, enabling a more balanced learning 
problem. We have found that, otherwise, the learning algorithm can have
a tendency to ignore the pose parameters and produce a smaller
diversity of target feature vectors. Finally, rather than a function 
of $\bf a$ and $\bf t$ alone, the decoder is a function 
of $\bf a$, $\bf p$, and $\bf t$. This again guarantees that the 
\emph{intermediate} representation is higher dimensional and facilitates
the learning of the decoder. We next discuss the details of the various
network modules.

\subsection{Network details} 

\noindent
\textbf{Encoder.}
The encoder consists of a pose and an appearance predictor.
The pose predictor implements the mapping ${\bf p} = {\cal P}({\bf x})$
from feature vectors $\bf x$ to pose parameters. The
poses are first internally mapped into a code vector ${\bf c} \in \mathbb{R}^N$
of dimensionality comparable to that of the appearance vector
$\bf a$. In the current implementation of FATTEN this is achieved
in three steps. First, the pose space is quantized into $N$ cells
of centroids ${\bf m}_i$. Each pose is then assigned to the cell
of the nearest representative ${\bf m}^*$ and represented by a 
$N$-dimensional one-hot encoding that identifies ${\bf m}^*$.
The pose mapping $\cal P$ is finally implemented
with a classifier that maps $\bf x$ into a vector of
posterior probabilities
\begin{equation}
  \mathbf{p} = [p({\bf m}_1|\mathbf{x}),\ldots,p({\bf m}_N|\mathbf{x})]
\end{equation}
on the $N-1$ probability simplex $\mathbb{P}^{N-1}$. 
This is implemented with a two-layer neural network, composed
of a fully-connected layer, batch normalization, and a ReLU, 
followed by a softmax layer. 

\vskip0.5ex
The appearance predictor implements the mapping 
${\bf a} = {\cal A}({\bf x})$ from feature vectors $\bf x$ to appearance 
descriptors $\bf a$. This is realized with a two-layer network,
where each layer consists of a fully-connected layer, batch normalization,
and a ELU layer. The outputs of the pose and appearance predictors
are concatenated with a one-hot encoding of the \emph{target} 
pose. Assuming that this pose belongs to the cell of centroid ${\bf m}_j$,
this is $\mathbf{t} = \mathbf{e}_j$, 
%\begin{equation}
%\end{equation}
where $\mathbf{e}_j$ is a vector of all zeros with a $1$ at position $j$.

\vskip1ex
\noindent
\textbf{Decoder.}
The decoder maps the vector of concatenated appearance and pose 
parameters\footnote{$\oplus$ denotes vector concatenation.} 
\begin{equation}
  [\mathbf{a}~\oplus~\mathbf{p}~\oplus~\mathbf{t}]
\end{equation}
into the residual $\hat{\bf x} - {\bf x}$. It is implemented 
with a two layer network, where the first layer contains
a sequence of fully-connected layer, batch normalization, and ELU,
and the second is a fully connected layer. The decoder output is
finally summed to the input feature vector $\bf x$ to produce
the target feature vector $\hat{\bf x}$.

\begin{figure*}[t]
\centering
\includegraphics[width=\textwidth]{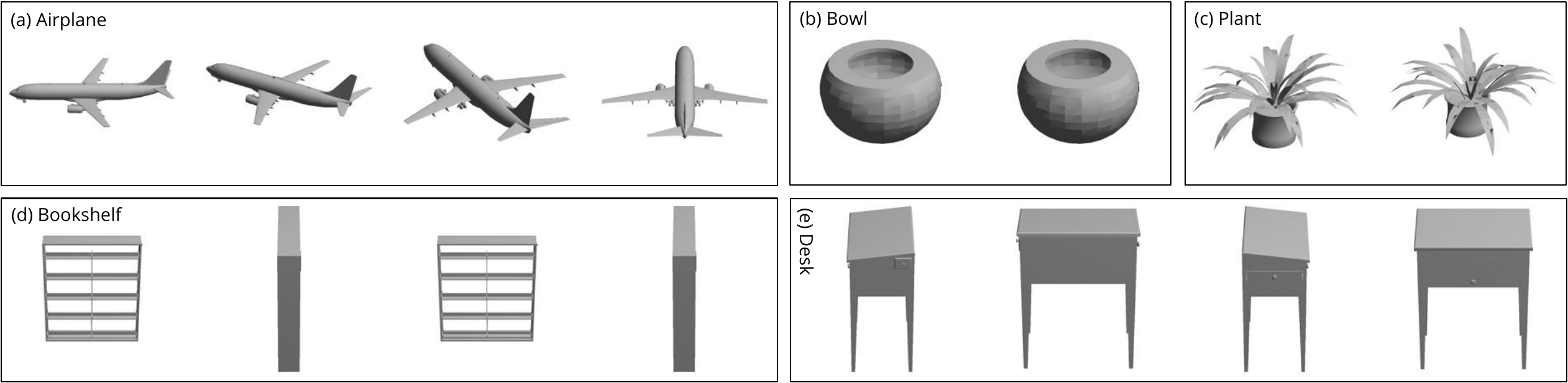}
%	\begin{center}
%		\begin{tabular}{cccccccc}
%			\includegraphics[width=0.08\linewidth]{figs/airplane_001.jpg} &
%			\includegraphics[width=0.08\linewidth]{figs/airplane_002.jpg} &
%			\includegraphics[width=0.08\linewidth]{figs/airplane_003.jpg} &
%			\includegraphics[width=0.08\linewidth]{figs/airplane_004.jpg} &
%			\includegraphics[width=0.08\linewidth]{figs/bowl_001.jpg} &
%			\includegraphics[width=0.08\linewidth]{figs/bowl_002.jpg} &
%			\includegraphics[width=0.08\linewidth]{figs/plant_001.jpg} &
%			\includegraphics[width=0.08\linewidth]{figs/plant_004.jpg} \\
%			(a.i) & (a.ii) & (a.iii) & (a.iv) & (b.i) & (b.ii) & (c.i) & (c.ii) \\
%			\includegraphics[width=0.08\linewidth]{figs/bookshelf_001.jpg} &
%			\includegraphics[width=0.08\linewidth]{figs/bookshelf_004.jpg} &
%			\includegraphics[width=0.08\linewidth]{figs/bookshelf_007.jpg} &
%			\includegraphics[width=0.08\linewidth]{figs/bookshelf_010.jpg} &
%			\includegraphics[width=0.08\linewidth]{figs/desk_001.jpg} &
%			\includegraphics[width=0.08\linewidth]{figs/desk_004.jpg} &
%			\includegraphics[width=0.08\linewidth]{figs/desk_007.jpg} &
%			\includegraphics[width=0.08\linewidth]{figs/desk_010.jpg} \\
%			(d.i) & (d.ii) & (d.iii) & (d.iv) & (e.i) & (e.ii) & (e.iii) & (e.iv) \\
%		\end{tabular}
%	\end{center}
	\caption{Exemplary ModelNet \cite{Wu15a} views: (a) Different views from one object (airplane);  
	(b)-(c) Symmetric object (bowl, plant) in different views; (d)-(e) Four views (bookshelf, desk) 
	with 90 degrees difference.}
	\label{fig:modelnet}
\end{figure*}

\subsection{Training}

The network is trained end-to-end, so as to optimize a multi-task loss
that accounts for two goals. The first goal is that the generated feature
vector $\hat{\bf x}$ indeed corresponds to the desired pose $\bf t$. This is 
measured by the pose loss, which is the cross-entropy loss
commonly used for classification, \ie,
\begin{equation}
  L_p(\mathbf{\hat{x}}, {\bf t}) = -\log \rho_j ({\cal P}(\mathbf{\hat{x}}))\enspace,
  \label{eq:Lp}
\end{equation}
where $\rho_j(v) = \frac{e^{v_j}}{\sum_k e^{v_k}}$ is the softmax function
and $j$ is the non-zero element of the one-hot vector ${\bf t} = {\bf e}_j$.
Note that, as shown in Fig.~\ref{fig:arch}, this requires passing
the target feature vector $\hat{\bf x}$ through the pose 
predictor ${\cal P}$.
It should be emphasized that this is only needed during training, albeit 
the loss in Eq.~\eqref{eq:Lp} can also be measured during inference, since the
target pose $\bf t$ is known. This can serve as a diagnostic of the 
performance of FATTEN.

\vskip0.5ex
The second goal is that the generated feature vector $\hat{\bf x}$
is assigned the same class label $y$ as the source vector ${\bf x}$.
This encourages the generation of features with high recognition
accuracy on the original object recognition problem. Recognition
accuracy depends on the 
network used to extract the feature vectors, denoted as CNN in 
Fig.~\ref{fig:arch}. Note that this network can be 
fine-tuned for operation with the FATTEN module in an end-to-end manner.
While FATTEN can, in principle, be applied to any such network, our 
implementation is based on the VGG16 model of \cite{Simonyan14a}. 
More specifically, we rely on the \texttt{fc7} activations of a fine-tuned 
VGG16 network as source and target features. The category predictor 
of Fig.~\ref{fig:arch} is then the \texttt{fc8} layer of this network.
The accuracy of this predictor is measured with a cross-entropy loss
\begin{equation}
  L_c(\mathbf{\hat{x}}, y) = -\log \rho_y (\mathbf{\hat{x}})\enspace,
  \label{eq:Lp}
\end{equation}
where $\rho(v)$ is the softmax output of this network. 
The \emph{multi-task loss} is then defined as 
\begin{equation}
  L(\mathbf{\hat{x}}, {\bf t}, y) = 
  L_a(\mathbf{\hat{x}}, {\bf t}) + L_c(\mathbf{\hat{x}}, y)\enspace.
\end{equation}

In general, it is beneficial to pre-train the pose predictor
${\cal P}({\bf x})$ and \emph{embed} it into the encoder-decoder structure.
This reduces the number of degrees of freedom the network, and minimizes
the ambiguity inherent to the fact that a given feature vector
could be consistent with multiple pairs of pose and appearance parameters.
For example, while all feature vectors $\bf x$ extracted from views of the 
same object should be constrained to map into the same appearance 
parameter value $\bf p$, we have so far felt no need
to enforce such constraint. This endows the network with robustness
to small variations of the appearance descriptor, due to occlusions,
etc. Furthermore, when 
a pre-trained pose predictor is used, \emph{only} the weights of the 
encoder/decoder need to be learned. The weights of the sub-networks 
used by the loss function(s) are fixed. This minimizes the chance that the 
FATTEN structure will over-fit to specific pose values or object categories.

\section{Experiments}
\label{section:experiments}

We first train and evaluate the FATTEN model on the artificial 
ModelNet~\cite{Wu15a} dataset (Sec.~\ref{subsection:modelnet}), and 
then assess its feature augmentation performance on the one-shot 
object recognition task introduced 
in~\cite{Dixit17a} (Sec.~\ref{subsection:oneshot}).

\subsection{ModelNet}
\label{subsection:modelnet}

\vskip1ex
\noindent
\textbf{Dataset.}
ModelNet~\cite{Wu15a} is a 3D artificial data set with 3D voxel grids. It 
contains $4000$ shapes from $40$ object categories. Given a 3D shape, 
it is possible to render 2D images from any pose. In our experiments, 
we follow the rendering strategy of~\cite{su2015multi}. $12$ virtual 
cameras are placed around the object, in increments of $30$ degrees along 
the $z$-axis, and $30$ degrees above the ground. Several rendered views are 
shown in Fig.~\ref{fig:modelnet}. The training and testing division is 
the same as in the ModelNet benchmark, using $80$ objects per category 
for training and $20$ for testing. However, the dataset contains some 
categories of symmetric objects, such as `bowl', which produce
identical images from all views (see Fig.~\ref{fig:modelnet}(b)) and
some that lack any distinctive information across views, such as `plant'
(see Fig.~\ref{fig:modelnet}(c)). For training, these objects are eliminated
and the remaining $28$ object categories are used.

%We choose to eliminate these objects and use 
%the remaining $28$ object categories for training. 
%After dataset cleaning, we finally obtain a dataset with view information as attribute labels, and do not have 
%any background noise.

%\begin{table*}
%	\begin{center}
%		\begin{tabular}{|l|c|}
%			\hline
%			Set & Category \\
%			\hline\hline
%			S(19)  & bathtub, bed, bookshelf, box, chair, counter, desk, door, dresser, garbage bin, \\
%			& lamp, monitor, night stand, pillow, sink, sofa, table, tv, toilet \\
%			\hline
%			T1(10) & picture, whiteboard, fridge, counter, books, stove, cabinet, printer, computer, ottoman \\
%			T2(10) & mug, telephone, bowl, bottle, scanner, microwave, coffee table, recycle, bin, cart, bench \\
%			\hline
%		\end{tabular}
%	\end{center}
%	\caption{List of categories in each set.}
%	\label{tab:oneshotsets}
%\end{table*}

\vskip1ex
\noindent
\textbf{Implementation.}
All feature vectors $\bf x$ are collected from the \texttt{fc7} activations 
of a fine-tuned VGG16 network. The pose predictor is trained with a 
learning rate $0.01$ for $1000$ epochs, and evaluated on the testing 
corpus. The complete FATTEN model is then trained for $10,000$ epochs with a 
learning rate of $0.01$. 
The angle range of $0^\circ$-$360^\circ$ are divided into $12$ non-overlapping intervals of size $30^\circ$ each, which are labeled as $0$-$11$. Any given angle value is then converted to a classification label based on the interval it belongs to.
%The whole 360-degree span of pose value is divided in to 12 non-overlapping 30-degree intervals, 
%which are labeled as $0$-$11$. Any given pose value is then converted to a classification label, based on the
%interval where it is.
%{\color{red}[TALK ABOUT THE POSE ENCODING]}

\subsubsection{Feature transfer results}

The feature transfer performance of FATTEN is assessed in two steps.
The accuracy of the pose predictor is evaluated first,
with the results listed in Table.~\ref{tab:pose}. The large majority
of the errors have magnitude of $180^\circ$. 
%We train and evaluate the performance of feature space transfer on the dataset discussed above.  First, results for attribute value %class prediction are shown in Table.~\ref{tab:pose}.  
This is not surprising, since ModelNet images have no texture. As
as shown in Fig.~\ref{fig:modelnet}(d)-(e), object views that differ by 
$180^\circ$ can be similar or even identical for some objects. However, 
this is not a substantial problem for transfer. Since two feature
vectors corresponding to the $180^\circ$ difference are close to each other 
in feature space, to the point where the loss cannot distinguish 
them clearly, FATTEN will generate target features close to the source,
which is the goal anyway. If these errors are disregarded, the pose
prediction has accuracy $90.8\%$.

\begin{table}
\begin{small}
\centering
\begin{tabular}{lccccccc}
  \toprule
  \textbf{Err.} [deg] & $0$ & $30$ & $60$ & $90$ & $120$ & $150$ & $180$ \\
  \midrule
  % Percentage & 72.26 & 2.23 & 1.07 & 4.02 & 0.88 & 1.00 & 18.54
  Perc. & $72.3$ & $2.2$ & $1.1$ & $4.0$ & $0.9$ & $1.0$ & $18.5$\\
  \bottomrule
\end{tabular}
\caption{Pose prediction error on ModelNet. \textit{Perc.} denotes the 
  percentage of error cases.}
	%	Pose predictor results on every error.}
	\label{tab:pose}
	\end{small}
\end{table}

\vskip0.5ex
The second evaluation step measures the feature transfer performance
of the whole network, given the pre-trained pose predictor.
During training, each feature in the training set is transferred to all 12 
views (including the identity mapping). During testing, this is
repeated for each test feature. 
%We train the transfer network for $10000$ epochs with a learning rate of $0.01$. 
The accuracy of the pose and category prediction of the features, generated on
the test corpus, is listed in Table~\ref{tab:transfer}. Note that, here,
category refers to object category or class. It is
clear that on a large synthetic dataset, such as ModelNet, FATTEN
can generate features of good quality, as indicated by the 
pose prediction accuracy of $96.2\%$ and the category prediction accuracy of 
$83.65\%$.

\begin{table}
\centering
\begin{tabular}{l|cc}
  \toprule
  & \textbf{Pose} & \textbf{Object category} \\
  \hline
  Accuracy [\%] & 96.20 & 83.65\\
  \bottomrule
\end{tabular}
\caption{Pose and category accuracy (in \%) of generated features, 
  on ModelNet. }
\label{tab:transfer}
\end{table}

\subsubsection{Retrieval with generated features}
\label{subsubsection:retrieval}

A set of retrieval experiments is performed on ModelNet to further
assess the effectiveness of FATTEN generated features. These experiments
address the question of whether the latter can be used to retrieve 
instances of (1) the same class or (2) the same pose. 
Since all features are extracted from the VGG16 \texttt{fc7} layer,
the Euclidean distance 
\begin{equation}
d_1(\mathbf{x},\mathbf{y}) = ||\mathbf{x} - \mathbf{y}||_2  
\end{equation}
is a sensible measure of similarity
between $\mathbf{x},$ and $\mathbf{y}$ for the purpose of
retrieving images of the same \emph{object category}. 
This is because the model is trained to map features with equal category 
labels to the same partitions of the feature space (enforced by the 
category loss $L_c$).
%A distance measure is needed in retrieval tasks. As we mainly work on features space (e.g. \texttt{FC7}), Euclidean distance is %commonly used as follows:
%\begin{equation} \label{eq:d1}
%d_1(\mathbf{x},\mathbf{y}) = ||\mathbf{x} - \mathbf{y}||_2
%\end{equation}
%, which will be the category distance in this task, since the feature extraction model is trained to map features with same category labels %together. 
However, $d_1$ is inadequate for \emph{pose} retrieval. Instead,
retrieval is based on the activation of the second fully-connected layer 
of the pose predictor ${\cal P}$, which is denoted $\gamma({\bf x})$. The 
\emph{pose distance function} is the defined as
\begin{equation}\label{eq:d2}
  d_2(\mathbf{x},\mathbf{y}) = ||\gamma(\mathbf{x}) - \gamma(\mathbf{y})||_2\enspace.
\end{equation}

%\begin{table}
%	\centering
%	\begin{tabular}{lccc}
%		\toprule
%		\textbf{Feature type}  & \textbf{(P)ose} & \textbf{(C)ategory} & \textbf{P} + \textbf{C} \\
%		\hline
%		Real 			& $54.58$ & \cellcolor{green!30}{$32.71$} & \cellcolor{green!30}{$23.65$} \\
%		Generated 	& \cellcolor{green!30}{$77.62$} & $28.89$ & $11.07$ \\
%		\bottomrule
%	\end{tabular}
%	\caption{Retrieval performance in mAP [\%] of real and generated 
%          features, on the testing portion ModelNet, for distance functions
%          $d_1, d_2$ and $d_c$, see Sec.~\ref{subsubsection:retrieval}.}
%	\label{tab:retrieval}
%\end{table}

\begin{table}
	\centering
	\begin{tabular}{lccc}
		\toprule
		\textbf{Feature type}  & \textbf{(P)ose} & \textbf{(C)ategory} & \textbf{P} + \textbf{C} \\
		\hline
		Real 			& $54.58$ & $32.71$ & $23.65$ \\
		Generated 	& $77.62$ & $28.89$ & $11.07$ \\
		\bottomrule
	\end{tabular}
	\caption{Retrieval performance in mAP [\%] of real and generated 
		features, on the testing portion ModelNet, for distance functions
		$d_1, d_2$ and $d_c$, see Sec.~\ref{subsubsection:retrieval}.}
	\label{tab:retrieval}
\end{table}

Finally, the performance of \emph{joint} category \& pose retrieval is measured
with a combined distance
\begin{equation}\label{eq:d}
  d_c(\mathbf{x},\mathbf{y}) = d_1(\mathbf{x},\mathbf{y}) + 
  \lambda d_2(\mathbf{x},\mathbf{y})\enspace.
\end{equation}
All queries and instances to be retrieved are based on \emph{generated} 
features from the \emph{testing} corpus of ModelNet.
%In experiments, all generated features in testing corpus are taken to retrieve instances in testing set. 
For each generated feature, three queries are performed: (1) 
\emph{Category}, (2) \emph{Pose}, and (3) \emph{Category \& Pose}. 
This is compared to the performance, on the same experiment, of the
real features extracted from the testing corpus. 
%Same setup is also applied on real features for comparison. 

\begin{figure*}[t!]
	\centering
	\includegraphics[width=0.9\linewidth]{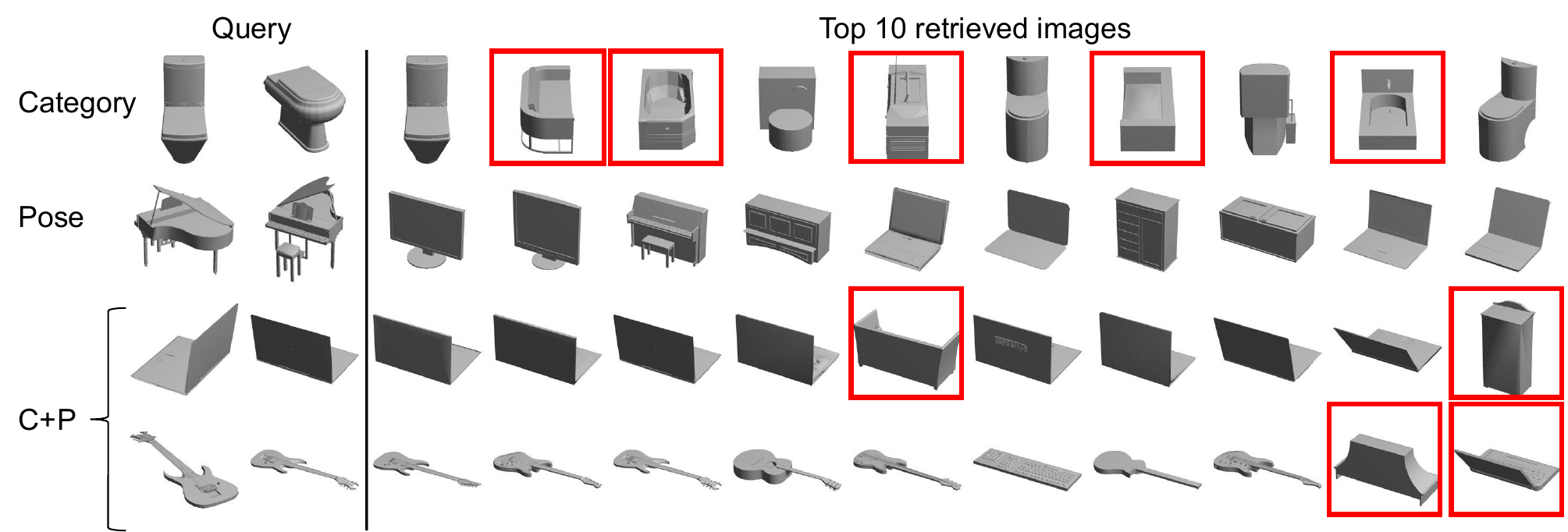}
	\caption{Some retrieval results for the experiments 
          of Sec.~\ref{subsubsection:retrieval}. The first two lines 
          refer to category and pose retrieval, lines 3-4 to category 
          retrieval and lines 5-6 to pose retrieval. Errors are highlighted 
          in \textcolor{red}{red}.  For each pair of figures in query part, the left one is original figure, while the right one is the real image corresponding to generated feature.
		\label{fig:retrieval}}
\end{figure*}

Retrieval results are listed in Table~\ref{tab:retrieval} and some 
retrieval examples are shown in Fig.~\ref{fig:retrieval}.
The generated features enable a very high mAP for \emph{pose retrieval}, 
even higher than the mAP of real features. This is strong evidence that 
FATTEN successfully encodes pose information in the transferred features. 
The mAP of the generated features on \emph{category retrieval} and the 
combination of both is comparatively low. However, the performance of real 
features is also weak on these tasks. This could be due to a failure of 
mapping features from the same category into well defined neighborhoods, 
or to the distance metric used for retrieval. While retrieval 
performs a nearest neighbor search under these metrics, the network
optimizes the cross-entropy loss on the softmax output(s) of both output 
branches of Fig.~\ref{fig:arch}. The distance of Eq.~\eqref{eq:d} may be a
particularly poor way to assess joint category and pose distances.
%This mismatch might also contribute to the mediocre performance on category
%retrieval. 
In the following section, we will see that using a strong classifier 
(\eg, a SVM) on the generated features produces significantly better
results.

%We ascribe this behavior to the failure of mapping features from same category together.

\subsection{One-shot object recognition}
\label{subsection:oneshot}
The experiments above provide no insight on whether FATTEN
generates meaningful features for tasks involving real world datasets. 
In this section, we assess feature transfer performance on a one-shot 
object recognition problem. On this task, feature transfer is used for 
for feature space ``fattening'' or \emph{data augmentation}.
The dataset and benchmark is collected from SUN-RGBD~\cite{Song15a}, 
following the setup of~\cite{Dixit17a}. 
%We will first describe the dataset briefly.

\vskip1ex
\noindent
\textbf{Dataset.}
The whole SUN-RGBD dataset contains $10335$ images and their corresponding depth maps. Additionally, 2D and 3D bounding boxes are available as ground truth for object detection. \emph{Depth} (distance from the camera plane) and 
\emph{Pose} (rotation around the vertical axis of the 3D coordinate system) 
are used as pose parameters in this task.
The depth range of $[0, 5)$ m is broken into
non-overlapping intervals of size 0.5m. An additional interval $[5, +\infty)$ is included for larger depth values.  
%Intervals with a span $0.5$ from $0$ to $5$ as well as $[5,+\infty)$ are chosen for \emph{Depth}. 
For pose, the angular range of $0^\circ$-$180^\circ$ is
divided into $12$ non-overlapping intervals of size $15^\circ$ each.
These intervals are used for one-hot encoding and system training. To allow a fair comparison with AGA, however, during testing, we restrict the desired pose $t$ to take the values $45^\circ$, $75^\circ$, ..., $180^\circ$ prescribed in~\cite{Dixit17a}. This is mainly to ensure that our system generates $11$ synthetic points along the Depth trajectory and $7$ along the \emph{Pose} trajectory similar to theirs.
 
%Degrees from $0^\circ$-$180^\circ$ with interval of $15^\circ$ ($12$ intervals) are used in \emph{Pose} for training, while $45^\circ$, $75^\circ$, ..., $180^\circ$ are selected for fairly comparison during testing. This involves $11$ target features for \emph{Depth}, and $7$ for \emph{Pose}.
% \textcolor{red}{[TALK ABOUT POSE/DEPTH ENCODING].}

The first $5335$ images of SUN-RGBD are used for training and the 
remaining $5000$ images for testing. However, if only ground truth bounding 
boxes are used for object extraction, the instances are neither balanced \wrt 
categories, nor \wrt pose/depth values. %if we only use ground truth bounding box to exact objects. 
To remedy this issue, a fast R-CNN~\cite{Girshick15a} object detector is 
fine-tuned on the dataset and the selective search proposals with 
IoU$>0.5$ (to ground truth boxes) and detection scores $>0.7$ are used to 
extract object images for training. As this strategy produces a sufficient 
amount of data, the training set can be easily balanced per category, as 
well as pose and depth. In the testing set, only ground truth bounding 
boxes are used to exact objects. All source features are exacted from the 
penultimate (\ie, \texttt{fc7}) layer of the fine-tuned fast R-CNN 
detector for all instances from both training and testing sets.

\vskip1ex
Evaluation is based on the source and target object classes defined 
in \cite{Dixit17a}. We denote $\mathcal{S}$ as
the source dataset, and let $\mathcal{T}_1$ and $\mathcal{T}_2$ denote two different (disjoint) target datasets; further, $\mathcal{T}_3=\mathcal{T}_1\cup \mathcal{T}_2$ denotes a third dataset that is a union of the first two. 
%\footnote{Also, $\mathcal{S} \cap \mathcal{T}_1 = \emptyset$ and
%$\mathcal{S} \cap \mathcal{T}_1 = \emptyset$.}
Table~\ref{tab:oneshotsets} lists all the object categories in each set. The instances in $\mathcal{S}$ are collected from the training portion
of SUN-RGBD only, while those in $\mathcal{T}_1$ and $\mathcal{T}_2$ are collected from the testing set. Further, $\mathcal{S}$ does not overlap with any $\mathcal{T}_i$ which ensurers that FATTEN has no access to shared knowledge 
between training/testing images or classes.

\vskip1ex
\noindent
\textbf{Implementation.}
The attribute predictors for \emph{pose} and \emph{depth} are trained with a learning rate of $0.01$ for $1000$ epochs. 
The feature transfer network is fine-tuned, starting from the weights obtained from the 
ModelNet experiment of Sec.~\ref{subsection:modelnet}, with a learning rate
of $0.001$ for $2000$ epochs. The classification problems on $\mathcal{T}_1$ and $\mathcal{T}_2$ are 10-class problems, whereas $\mathcal{T}_3$ is a 
20-class problem, respectively. As a \emph{baseline} for one-shot learning, we train a linear SVM using \emph{only} a single instance per class. We then feed those same instances into the feature transfer network to generate artificial features for different values of depth and pose. Specifically, we use $11$ different values for depth and $7$ for pose. After feature synthesis, a linear 
SVM is trained with the same parameters on the now \emph{augmented} (``fattened'') feature set (source and target features). 

\begin{table}
\centering
\begin{small}
\begin{tabular}{ll|ll}
\toprule
\multicolumn{2}{c}{$\mathcal{S}$ (19, Source)} & $\mathcal{T}_1$ (10) & $\mathcal{T}_2$ (10) \\
\midrule 
bathtub 	& lamp 		& picture 		& mug \\
bed 		& monitor 	& whiteboard		& telephone \\
bookshelf	& night stand & fridge 		& bowl \\  
box			& pillow	& counter 		& bottle \\
chair		& sink		& books 			& scanner \\
counter		& sofa		& stove			& microwave \\
desk		& table		& cabinet 		& coffee table\\
door		& tv		& printer 		& recycle bin \\
dresser		& toilet	&  computer		& cart \\
garbage bin	& 			& ottoman 		& bench \\
\bottomrule
%lamp		&			&  				& \\
%monitor		&& & \\
%night stand && & \\
%pillow	 	&& & \\
%sink		&& & \\
%sofa		&& & \\
%table		&& & \\
%tv			&& & \\
%toilet 		&& & \\
\end{tabular}
\end{small}
\caption{List of object categories in the source $\mathcal{S}$ training
set and the two target/evaluation sets $\mathcal{T}_1$ and $\mathcal{T}_2$.
\label{tab:oneshotsets}}
\end{table}

\subsubsection{Results}

Table~\ref{tab:oneshotresult} lists the averaged one-shot 
recognition accuracies (over $500$ random repetitions) for 
all three evaluation sets $\mathcal{T}_i$
%We repeat this process for $500$ 
%times and report the average recognition accuracy.
For comparison, five-shot results in the same augmentation setup are also reported. Table~\ref{tab:oneshotresult} additionally lists
the recognition accuracies of two recently proposed strategies to data augmentation, 
\ie, feature hallucination as introduced in  \cite{hariharan2016low} 
as well as attribute-guided augmentation (AGA) of  \cite{Dixit17a}. 

\vskip0.5ex
Table~\ref{tab:oneshotresult} supports some conclusions. \emph{First}, when compared to the SVM baseline, FATTEN achieves a 
remarkable and consistent improvement  of around $10$ percentage points
on all evaluation sets. This indicates that FATTEN can actually 
embed the pose information into features and effectively ``fatten''
the data used to train the linear SVM classifier. 
\emph{Second}, and most notably, FATTEN achieves a significant improvement 
(about $5$ percentage points) over AGA, and an even larger improvement 
over the feature hallucination approach of \cite{hariharan2016low}.
The improved performances of FATTEN over AGA and AGA over hallucination
show that it is important 1) to exploit the structure of the pose 
manifold (which only FATTEN and AGA do), and 2) to rely on models that
can capture defining properties of this manifold, such as continuity and
smoothness of feature trajectories (which AGA does not).

% While FATTEN has conceptual similarities to AGA, we attribute its improved performance 
% to the fact that AGA heavily relies on the performance of the attribute value predictor to select \emph{which} transfer
% function to use. 
% In cases where this prediction fails, the ``wrong'' synthesis function will be chosen and, consequently, the generated feature(s) are 
% suboptimal. FATTEN circumvents this
% drawback, as only one feature transfer network needs to be trained and the synthesis strategy
% is handled implicitly.

\vskip0.5ex
While the feature hallucination strategy works remarkably well in the 
ImageNet1k low-shot setup used in \cite{hariharan2016low}, 
Table~\ref{tab:oneshotresult} only shows marginal gains over
the baseline (especially in the one-shot case). 
There may be several reasons as to why it fails in this setup. 
First, the number of examples per category ($k$ in the 
notation of  \cite{hariharan2016low}) is a hyper-parameter 
set through cross-validation. To make the comparison fair, we chose to 
use the same value in all methods, which is $k=19$. This may not be the 
optimal setting for \cite{hariharan2016low}. Second, we adopt the same 
number of clusters as used by the authors when training the generator. 
However, the best value may depend on the dataset 
(ImageNet1k in \cite{hariharan2016low} \vs SUN-RGBD here). Without 
clear guidelines of how to set this parameter, it seems challenging 
to adjust it appropriately. Third, all results of \cite{hariharan2016low} 
list the top-5 accuracy, while we use top-1 accuracy. Finally, FATTEN 
takes advantage of pose and depth to generate more features, while the 
hallucination feature generator is \emph{non-parametric} and does not 
explicitly use this information for synthesis.

\vskip0.5ex
The improvement of FATTEN over AGA can most likely be attributed
to 1) the fact that AGA uses separate synthesis functions (trained independently) 
and 2) failure cases of the pose/depth predictor that determines \emph{which}
particular synthesis function is used. In case of the latter, generated features
are likely to be less informative, or might even confound any subsequent
classifier.

\begin{table}
\centering
		\begin{tabular}{lcccc}
			\toprule
			
			 & \textbf{Baseline} & \textbf{Hal.} \cite{hariharan2016low} & \textbf{AGA} \cite{Dixit17a} & \textbf{FATTEN} \\
			\midrule
			& \multicolumn{4}{c}{\emph{One-shot}} \\
			\midrule
			$\mathcal{T}_1$ (10) & 33.74 & 35.43 & 39.10 & 44.99 \\
			$\mathcal{T}_2$ (10)  & 23.76 & 21.12 & 30.12 & 34.70 \\
			$\mathcal{T}_3$ (20)  & 22.84 & 21.67 & 26.67 & 32.20 \\
			\midrule
			& \multicolumn{4}{c}{\emph{Five-shot}} \\
			\midrule
			$\mathcal{T}_1$ (10) & 50.03 & 50.31 & 56.92 & 58.82 \\
			$\mathcal{T}_2$ (10)  & 36.76 & 38.07 & 47.04 & 50.69 \\
			$\mathcal{T}_3$ (20)  & 37.37 & 38.24 & 42.87 & 47.07 \\
			\bottomrule
		\end{tabular}
	\caption{One-shot and five-shot recognition accuracy for three different few-shot recognition problems, constructed
	from the SUN-RGBD dataset. The recognition accuracies (in \%) are averaged over 500 random repetitions of the experiment. 
	The \textbf{Baseline} denotes the recognition accuracy achieved by a linear SVM, trained on \emph{single} 
	instances of each class only.}
	\label{tab:oneshotresult}
\end{table}

%for two reasons:
%meaningful gain compared to baseline mainly because of two reasons: 
%(1) the generation model doesn't take advantage of attributes and (2) it has several hyper-parameters which 
%need to be carefully adjusted during validation; this, however, is not applicable in our dataset.
%
%I THINK WE NEED A STRONGER ARGUMENT HERE FOR WHY HALLUCINATION DOES NOT WORK 
%WELL IN OUR SETUP - LIMITATIONS? IS HYPERPARAMETER TUNING NOT POSSIBLE AT ALL IN OUR
%CASE?

\section{Discussion}
\label{section:discussion}

The proposed architecture to data augmentation in feature space, FATTEN, aims  
to learn trajectories of feature responses, induced by variations in image properties 
(such as pose). These trajectories can then be easily
traversed via \emph{one} learned mapping function which, when applied to instances of 
novel classes, effectively enriches the feature space by additional samples corresponding to 
a desired change, \eg, in pose. This ``fattening'' of the feature space is highly beneficial in situations 
where the collection of large amounts of adequate training data to cover these variations
would be time-consuming, if not impossible. In principle, FATTEN can be used for any kind 
of desired (continuous) variation, so long as the trajectories can be learned from an 
external dataset. By discretizing the space of variations, \eg, the rotation angle in 
case of pose, we also effectively reduce the dimensionality of the learning problem
and ensure that the approach scales favorably w.r.t. different resolutions of desired changes.
Finally, it is worth pointing out that feature space transfer via FATTEN is not limited to 
object images; rather, it is a generic architecture in the sense that any variation could, in principle, 
be learned and transferred.

{\small
\bibliographystyle{ieee}
\bibliography{halucination,egbib}

\begin{thebibliography}{10}\itemsep=-1pt

\bibitem{Belkin03a}
M.~Belkin and P.~Niyogi.
\newblock Laplacian eigenmaps for dimensionality reduction and data
  representation.
\newblock {\em Neural Computation}, 15(6):1373--1396, 2003.

\bibitem{Deng09a}
J.~Deng, W.~Dong, R.~S. L.-J. Li, K.~Li, and L.~Fei-Fei.
\newblock Imagenet: A large-scale hierarchical image database.
\newblock In {\em CVPR}, 2009.

\bibitem{Dixit17a}
M.~Dixit, R.~Kwitt, M.~.Niethammer, and N.~Vasconcelos.
\newblock Aga: Attribute-guided augmentation.
\newblock In {\em CVPR}, 2017.

\bibitem{Flin17a}
C.~Finn, P.~Abbeel, and S.~Levine.
\newblock Model-agnostic meta-learning for fast adaptation of deep networks.
\newblock {\em CoRR}, abs/1703.03400, 2017.

\bibitem{Gatys16a}
L.~A. Gatys, A.~S. Ecker, and M.~Bethge.
\newblock Image style transfer using convolutional neural networks.
\newblock In {\em CVPR}, 2016.

\bibitem{Girshick15a}
R.~Girshick.
\newblock Fast {R-CNN}.
\newblock In {\em ICCV}, 2015.

\bibitem{Goodfellow14a}
I.~Goodfellow, J.~Pouget-Abadie, M.~Mirza, B.~Xu, D.~Warde-Farley, S.~Ozair,
  A.~Courville, and Y.~Bengio.
\newblock Generative adversarial nets.
\newblock In {\em NIPS}, 2014.

\bibitem{hariharan2016low}
B.~Hariharan and R.~Girshick.
\newblock Low-shot visual recognition by shrinking and hallucinating features.
\newblock {\em CoRR}, abs/1606.02819, 2016.

\bibitem{He16a}
K.~He, X.~Zhang, S.Ren, and J.~Sun.
\newblock Deep residual learning for image recognition.
\newblock In {\em CVPR}, 2016.

\bibitem{Isola17a}
P.~Isola, J.~Zhu, T.~Zhou, and A.~Efros.
\newblock Image-to-image translation with conditional adversarial networks.
\newblock In {\em CVPR}, 2017.

\bibitem{Kalogerakis16a}
E.~Kalogerakis, M.~Averkiou, S.~Maji, and S.~Chaudhuri.
\newblock 3d shape segmentation with projective convolutional networks.
\newblock {\em CoRR}, abs/1612.02808, 2016.

\bibitem{Karphaty15a}
A.~Karpathy and L.~Fei-Fei.
\newblock Deep visual-semantic alignments for generating image descriptions.
\newblock In {\em CVPR}, 2015.

\bibitem{knopp2010hough}
J.~Knopp, M.~Prasad, G.~Willems, R.~Timofte, and L.~Van~Gool.
\newblock Hough transform and 3d surf for robust three dimensional
  classification.
\newblock {\em Computer vision--ECCV 2010}, pages 589--602, 2010.

\bibitem{Krizhevsky12a}
A.~Krizhevsky, I.~Sutskever, and G.~E. Hinton.
\newblock {Imagenet} classification with deep convolutional neural networks.
\newblock In {\em NIPS}, 2012.

\bibitem{Lampert14a}
C.~H. Lampert, H.~Nickisch, and S.~Harmeling.
\newblock Attribute-based classification for zero-shot visual object
  categorization.
\newblock {\em TPAMI}, 36(3):453--465, 2014.

\bibitem{LeCun04a}
Y.~LeCun, F.~Huang, and L.~Bottou.
\newblock Learning methods for generic object recognition with invariance to
  pose and lighting.
\newblock In {\em CVPR}, 2004.

\bibitem{Nene96a}
S.~Nene, S.~Nayar, and H.~Murase.
\newblock Columbia object image library.
\newblock Technical Report CUCS-006-96, Columbia University, 1996.

\bibitem{Pathak16a}
D.~Pathak, P.~Kr\"ahenb\"uhl, J.~Donahue, T.~Darrell, and A.~Efros.
\newblock Context encoders: Feature learning by inpainting.
\newblock In {\em CVPR}, 2016.

\bibitem{Peng15a}
X.~Peng, B.~Sun, K.~Ali, and K.~Saenko.
\newblock Learning deep object detectors from 3d models.
\newblock In {\em ICCV}, 2015.

\bibitem{Qi16a}
C.~R. Qi, H.~Su, K.~Mo, and L.~J. Guibas.
\newblock Pointnet: Deep learning on point sets for 3d classification and
  segmentation.
\newblock {\em CoRR}, abs/1612.00593, 2016.

\bibitem{Qi16b}
C.~R. Qi, H.~Su, M.~Nie{\ss}ner, A.~Dai, M.~Yan, and L.~J. Guibas.
\newblock Volumetric and multi-view cnns for object classification on 3d data.
\newblock {\em CoRR}, abs/1604.03265, 2016.

\bibitem{Ravi17a}
S.~Ravi and H.~Larochelle.
\newblock Optimization as a model for few-shot learning.
\newblock In {\em ICLR}, 2017.

\bibitem{Ren15a}
S.~Ren, K.~He, R.~Girshick, and J.~Sun.
\newblock Faster {R-CNN}: Towards real-time object detection.
\newblock In {\em NIPS}, 2015.

\bibitem{RomeraParedes15a}
B.~Romera-Paredes and P.~Torr.
\newblock An embarrassingly simple approach to zero-shot learning.
\newblock In {\em ICML}, 2015.

\bibitem{Roweis00a}
S.~T. Roweis and L.~K. Saul.
\newblock Nonlinear dimensionality reduction by locally linear embedding.
\newblock {\em Science}, 290(5500):2323--2326, 2000.

\bibitem{Santoro16a}
A.~Santoro, S.~Bartunov, M.~Botvinick, D.~Wierstra, and T.~Lillicrap.
\newblock Meta-learning with memory-augmented neural networks.
\newblock In {\em ICML}, 2016.

\bibitem{Simonyan14a}
K.~Simonyan and A.~Zisserman.
\newblock Very deep convolutional networks for large-scale image recognition.
\newblock {\em CoRR}, abs/1409.1556, 2014.

\bibitem{Socher13a}
R.~Socher, M.~Ganjoo, C.~D. Manning, and A.~Y. Ng.
\newblock Zero-shot learning through cross-modal transfer.
\newblock In {\em NIPS}, 2013.

\bibitem{Song15a}
S.~Song, S.~Lichtenberg, and J.~Xiao.
\newblock {SUN RGB-D}: A {RGB-D} scene understanding benchmark suite.
\newblock In {\em CVPR}, 2015.

\bibitem{su2015multi}
H.~Su, S.~Maji, E.~Kalogerakis, and E.~Learned-Miller.
\newblock Multi-view convolutional neural networks for 3d shape recognition.
\newblock In {\em ICCV}, 2015.

\bibitem{Su15a}
H.~Su, C.~Qi, Y.~Li, and L.~Guibas.
\newblock Render for {CNN}: Viewpoint estimation in images using cnns trained
  with rendered 3d model views.
\newblock In {\em ICCV}, 2015.

\bibitem{Szegedy15a}
C.~Szegedy, W.~Liu, Y.~Jia, P.~Sermanet, S.~Reed, D.~Anguelov, D.~Erhan,
  V.~Vanhoucke, and A.~Rabinovich.
\newblock Going deeper with convolutions.
\newblock In {\em CVPR}, 2015.

\bibitem{Tang10a}
K.~Tang, M.~Tappen, R.~Sukthankar, and C.~Lampert.
\newblock Optimizing one-shot recognition with micro-set learning.
\newblock In {\em CVPR}, 2010.

\bibitem{vdM08a}
L.~{van der Maaten} and G.~Hinton.
\newblock Visualizing high-dimensional data using t-sne.
\newblock {\em Journal of Machine Learning Research}, 9:2579--2605, 2008.

\bibitem{Vinyals15a}
O.~Vinjays, A.~Toshev, S.~Bengio, and D.~Erhan.
\newblock Show and tell: A neural image caption generator.
\newblock In {\em CVPR}, 2015.

\bibitem{Vinyals16a}
O.~Vinyals, C.~Blundell, T.~Lillicrap, k.~kavukcuoglu, and D.~Wierstra.
\newblock Matching networks for one shot learning.
\newblock In {\em NIPS}, 2016.

\bibitem{Wu15a}
Z.~Wu, S.~Song, A.~Khosla, F.~Yu, L.~Zhang, X.~Tang, and J.~Xiao.
\newblock {3D ShapeNets}: A deep representation for volumetric shape modeling.
\newblock In {\em CVPR}, 2015.

\end{thebibliography}
}

\end{document}